# Fuzzy Convolution Neural Networks for Tabular Data Classification

**Arun D. Kulkarni**
**Computer Science Department, University of Texas at Tyler, Tyler, TX 75799, USA**

**Corresponding author: Arun D. Kulkarni (e-mail: akulkarni@uttyler.edu).**

**ABSTRACT** Recently, convolution neural networks (CNNs) have attracted a great deal of attention due to their remarkable performance in various domains, particularly in image and text classification tasks. However, their application to tabular data classification remains underexplored. There are many fields such as bioinformatics, finance, medicine where non-image data are prevalent. Adaption of CNNs to classify non-image data remains highly challenging. This paper investigates the efficacy of CNNs for tabular data classification, aiming to bridge the gap between traditional machine learning approaches and deep learning techniques. We propose a novel framework fuzzy convolution neural network (FCNN) tailored specifically for tabular data to capture local patterns within feature vectors. In our approach, we map feature values to fuzzy memberships. The fuzzy membership vectors are converted into images that are used to train CNN model. The trained CNN model is used to classify unknown feature vectors. To validate our approach, we generated six complex noisy data sets. We used randomly selected seventy percent samples from each data set for training and thirty percent for testing. The data sets were also classified using the state-of-the-art machine learning algorithms such as the decision tree (DT), support vector machine (SVM), fuzzy neural network (FNN), Bayes' classifier, and Random Forest (RF). Experimental results demonstrate that our proposed model can effectively learn meaningful representations from tabular data, achieving competitive or superior performance compared to existing methods. Overall, our finding suggests that the proposed FCNN model holds promise as a viable alternative for tabular data classification tasks, offering a fresh prospective and potentially unlocking new opportunities for leveraging deep learning in structured data analysis.

**INDEX TERMS** Fuzzy Logic, Convolution Neural Networks, Deep Learning, Tabular data, Machine Learning, Classification

## I. INTRODUCTION

In the era of data-driven decision-making, the ability to accurately classify and analyze tabular data plays a crucial role across various domains, including finance, healthcare, marketing, and beyond. Traditionally, this task has been approached using machine learning algorithms such as decision trees, support vector machines, random forests, and artificial neural network models which rely on handcrafted features and explicit rule-based representations. However, with the advent of deep learning, particularly Convolutional Neural Networks (CNNs), there has been a change in thinking in how complex patterns and relationships in data can be learned directly from raw inputs. CNN models are conventionally used for image classification due to their high performance, availability of various architectures, and availability of graphical processing units (GPUs). CNN models excel in achieving high accuracy for image data classification. While CNNs have demonstrated remarkable success in tasks like image and text classification, their application to tabular data classification has received comparatively less attention. Tabular data typically consist of rows and columns, where each column represents a feature. CNNs are designed for processing grid-like data to capture spatial dependencies in data like images, where relationships exist both horizontally and vertically. However, tabular data do not possess the same grid-like structure as images, and the relationships between features are not spatial in nature. CNNs offer several advantages over traditional machine learning techniques. Firstly, they provide flexibility and support iterative learning. Secondly, deep networks enable the generation of tabular data, which can help alleviate class imbalance issues. Thirdly, neural networks can be employed for multimodal learning problems, where tabular data serve as one of many input modalities [1]. CNNs exploit the spatial locality of features in data, which is not applicable to tabular data. In tabular data, the relationships between features are not spatially organized but rather depend on their interdependencies. CNNs excel at capturing local patterns in data due to their convolutional kernels, which slide across the input. Convolutional kernels are excellent feature extractors that exploit two properties of the input images: local connectivity and spatial locality. Local connectivity means that each kernel is connected to a small region of the input image when performing the convolution. The spatial locality property means that the pixels where the convolutional kernel is applied are highly correlated, and usually processing them jointly makes it possible to extract meaningful feature representations. For example, a single convolutional kernel can learn to extract edges, textures, shapes, and gradients. While this is effective for tasks like image classification where local features matter, tabular data often require capturing both local and global patterns to make accurate predictions. Fully connected neural networks or tree-based models can better capture these global patterns. In tabular data, where the number of features can be relatively small compared to other domains like images or text, the efficiency of CNNs might not be fully utilized. CNNs require a large amount of data to effectively learn the parameters of the convolutional filters. Tabular datasets are often smaller compared to image datasets, making it challenging for CNNs to generalize well. CNNs are known for their black-box nature, making it challenging to interpret how they make predictions, especially in the context of tabular data where interpretability is often crucial for understanding model decisions and gaining insights from the data. Despite these challenges, CNNs offer the potential to automatically learn hierarchical representations of tabular data, capturing both local and global patterns within feature vectors. There have been attempts to adapt CNNs for tabular data. One of the approaches to classifying tabular data is to transform the tabular data into images. CNNs require fixed-size input tensors, typically with three dimensions (width, height, channels). Tabular data, on the other hand, can have varying numbers of features, and the order of features may not have any significance. The effectiveness of CNNs on tasks involving image processing is because they consider the spatial structure of data, capturing spatially local input patterns. In tabular data, the relationships between features are often more complex and might not be easily captured by 1-D convolutions alone. We cannot feed a tabular dataset straight forward to a convolutional layer because tabular features are not spatially correlated [2]. Most tabular data do not assume a spatial relationship between features, and thus are unsuitable for modeling using CNNs. CNNs are designed to automatically learn hierarchical representations of features in data. In tabular data, the importance of features and their relationships may not follow a hierarchical structure, making it less suitable for CNNs. CNN models encounter challenges such as the vanishing gradient problem, which is mitigated by employing the entropy loss function with linear rectified units (ReLu) in the output layer. Another issue is overfitting, especially prevalent in small datasets. Both Alex Net and ResNet-50 employ several techniques to mitigate overfitting. Alex Net uses techniques such as data augmentation, drop out, and weight decay, while ResNet-50 uses techniques such as data augmentation, batch normalization, global average pooling, and weight decay.

Tabular data, characterized by structured rows and columns, present unique challenges such as dealing with heterogeneous feature types and capturing interactions between features.

This paper aims to explore the efficacy of CNNs for tabular data classification, filling a crucial gap in the literature and advancing the understanding of deep learning methods in structured data analysis. We propose a novel Fuzzy Convolution Neural Network (FCNN) architecture specifically tailored for tabular data. We introduce a method for mapping a feature vector onto an image. This mapping involves assigning feature values to their corresponding fuzzy membership values. We employed fuzzy membership functions representing five term sets: *very_low*, *low*, *medium*, *high*, and *very_high*. We map each fuzzified feature vector into an image. In our earlier research work, we converted feature vectors into images by mapping features and their ratios to rectangular shapes in the image canvas [3]. In this research work, we assign features to their fuzzy membership values and represent fuzzy membership values by square shapes within the mapped image.

Through extensive experimentation on six complex datasets, we evaluate the performance of our proposed model FCNN against traditional machine learning algorithms and the fuzzy neural network model. The contributions of this paper are a) we introduce a novel FCNN architecture designed for tabular data classification, addressing the unique challenges associated with structured data. b) we conducted comprehensive experiments to demonstrate the effectiveness of FCNNs in comparison to conventional machine learning approaches for tabular data classification tasks and demonstrated that the proposed FCNN model performs equal or superior to the state-of-the-art methods. The paper's structure is as follows: Section II delves into related work, Section III presents the framework for the FCNN model, Section IV covers implementation and results, and Section V presents conclusions and future work.

## II. RELATED WORK

Various machine learning algorithms are employed for the classification of tabular data. These include the minimum distance classifier, the maximum likelihood classifier (MLC), and non-parametric techniques such as the support vector machine (SVM), decision tree (DT), ensemble of decision trees, multi-layer perceptron model, fuzzy inference system, and fuzzy neural networks. The maximum likelihood classification algorithm assumes a normal distribution for feature values, computing the mean vector and covariance matrix for each class using training set data. By applying Bayes' rule, the classifier calculates posterior probabilities and assigns the sample to the class with the highest posterior probability [4].

Decision tree (DT) classifiers are non-parametric classifiers that do not require any priori statistical assumptions regarding the distribution of data. The structure of a decision tree consists of a root node, non-terminal nodes, and terminal nodes. The data are recursively divided down the decision tree according to the defined classifier framework. One of the most popular algorithms for constructing a decision tree is the ID3 algorithm. The ID3 induction tree algorithm has proven to be effective when working with large datasets that have several features, where it is inefficient for human experts to process. C4.5 is a supervised learning algorithm that is a descendant of the ID3 algorithm. C4.5 allows the usage of both continuous and discrete attributes. The main problem with decision trees is overfitting [5, 6]. Random Forest (RF) is based on tree classifiers. It implements several classification trees. The input vector is classified with each tree in the forest. Each tree provides a classification or "votes" for that class. The RF then selects the classification with the most votes among all the trees. The main advantages of Random Forest are unparalleled accuracy among current algorithms, efficient implementation on large datasets, and an easily saved structure for future use of pre-generated trees. In ensemble, results of trained classifiers are combined through a voting process. The most widely used ensemble methods are boosting and bagging [7]. Support vector machines (SVMs) are supervised non-parametric statistical learning methods. SVMs aim to find a hyperplane that separates training samples into a predefined number of classes. Vapnik [8] proposed the SVM algorithm. Operating as a binary classifier, SVM assigns a sample to one of the two linearly separable classes. In this algorithm, two hyperplanes are chosen to not only maximize the distance between the two classes but also to exclude any points between them. Nonlinearly separable classes are accommodated by extending the SVM algorithm to map samples into a higher-dimensional feature space. The SVM algorithm is especially well-suited for tabular data due to its adeptness in handling small datasets, frequently yielding higher classification accuracy compared to traditional methods.

Neural networks are favored for classification due to their parallel processing capabilities, as well as their learning and decision-making prowess. Several studies have aimed to evaluate neural networks' performance compared to traditional statistical methods for tabular data. Neural networks equipped with learning algorithms like backpropagation (BP) can extract insights from training samples and are utilized in tabular data analysis [9]. With advancements in hardware and algorithms, neural networks (NNs) have evolved into deep neural networks (DNNs) and convolutional neural networks (CNNs). CNNs stand out as highly effective learning algorithms for understanding image content and have displayed remarkable performance in various computer vision tasks. CNN models employ multiple layers of nonlinear information processing units. The machine learning community's interest in CNN surged after the ImageNet competition in 2012, where Alex Net achieved record-breaking results in classifying images from a dataset containing over 1.2 million images spanning one thousand classes [10]. Alex Net was built upon principles utilized in LeNet. Deep Convolutional Neural Networks (DCNNs) have heralded breakthroughs in processing images, videos, speech, and audio [11]. CNN models consist of convolution and pooling layers organized followed by one or more fully connected layers. They operate as feed-forward networks. In convolution layers, inputs undergo convolution with a weighted kernel, and the output is then passed through a nonlinear activation function to the subsequent layer. The primary aim of the pooling layer is to reduce spatial resolution. Rawat and Wang [12] offer a comprehensive survey of CNNs. Zhang et al. [13] present a taxonomy of CNN models. CNNs have the capability to learn internal representations directly from raw pixels and are hierarchical learning models capable of feature extraction [14]. Khan et al. [15] in their review article, categorized DCNN architectures into seven groups. Deep learning enables computational models composed of multiple processing layers to learn representations of data with various levels of abstraction. Recent advancements in CNN models have been facilitated by the accessibility of fast graphical processing units (GPUs) and the availability of extensive datasets.

The primary advantage of CNNs is their capacity to learn from input data and make decisions. However, due to the substantial number of parameters, they can be challenging to interpret and are often regarded as black boxes since they do not transparently explain how outcomes are achieved. Fuzzy Logic (FL) systems, on the other hand, excel at explaining their decisions but struggle with learning from input data. Combining FL and CNN can mitigate the drawbacks of each approach to create a more robust and flexible computational system. Talpur et al. [16] in their survey article, detail methods for integrating FL and DNN to create hybrid systems. One approach involves a sequential structure, where fuzzy systems and DNN operate sequentially. In this structure, there are two possibilities: a) converting input data into fuzzy sets, followed by processing the fuzzified data with the DNN, and b) the DNN model aiding the fuzzy system in determining desired parameters. Another method to combine FL and CNN is by utilizing CNN for feature extraction, transforming the output of the final convolution layer for fuzzy classification Sarabakha et al. [17] propose a DFNN structure where input features are fed to the fuzzification layer, and the fuzzified vector serves as input to fully connected hidden layers. Fuzzy

Deep Neural Networks (FDNNs) have been employed in many practical applications. FDNNs represent a compelling constructive collaboration between fuzzy logic and neural networks, offering a powerful tool for managing uncertainty and intricate relationships in real-world applications. Das et al. [18] provide a survey of FDNN systems.

Deep Convolutional Neural Network (DCNN) models have demonstrated remarkable performance and have consequently been widely adopted for computer vision tasks. However, adapting them to tabular data remains highly challenging. Sun et al. [19] introduced a method called SuperTML to convert tabular data into images. The algorithm adopts the concept of the Super Characters method for addressing machine learning tasks with tabular data. Initially, the input tabular features are projected onto a two-dimensional embedding and then fed into fine-tuned two-dimensional CNN models for classification. They validated the algorithm using four datasets, and experimental results demonstrate that SuperTML method achieves state-of-the-art results on both large and small tabular datasets. The main difference between SuperTML and FCNN method is that SuperTML method is based on success of Super Characters method in text classification. Whereas FCNN method maps fuzzy membership values into rectangular shapes. FCNN method is based on shape recognition. Sharma et al. [20] developed a method called DeepInsight to convert non-image data into images suitable for CNNs. Their approach constructs the image by grouping similar features together and positioning dissimilar ones farther apart, facilitating the collective utilization of neighboring elements. They evaluated their algorithm using four distinct datasets and compared the results against state-of-the-art classifiers such as decision trees, AdaBoost, and Random Forest. Their model exhibited superior classification accuracy across all datasets. In DeepInsight method feature vectors are transformed into feature matrices that are represented by pixels in the mapped image. The method is more suitable for large data sets. Zhu et al. [21] proposed a method named Image Generator for Tabular Data (IGTD) to convert tabular data into images by assigning features to pixel positions in a way that similar features are placed close to each other. The algorithm assigns each feature to a pixel in the image, generating an image for each data sample where the pixel intensity corresponds to the value of the respective feature in the sample. The algorithm seeks to optimize the assignment of features to pixels by minimizing the difference between the ranking of pairwise distances between features and the ranking of pairwise distances between assigned pixels. They applied the algorithm to two datasets. Their results demonstrate that CNNs trained on IGTD images yield the highest average prediction performance in cross-validation on both datasets. Du et al. [22] Have proposed a neural network architecture TabularNet to simultaneously extract spatial and relational information from tables. The spatial encoder of the TabularNet utilizes the row/column level pooling and the Bidirectional Gated Recurrent Unit (Bi-GRU) to capture statistical information and local positional correlation, respectively. Their experiments show that TabularNet significantly outperforms the state-of-the-art ML algorithms. Arik et al. [23] propose a high-performance and interpretable deep tabular data learning architecture called TabNet that uses sequential attention to choose which features to reason from at each decision step, enabling interpretability and more efficient learning. Besides robust performance, TabNet provides explainable insights on its reasoning, both locally and globally. Borisov et al. [1] provide an overview of deep learning methods tailored for tabular data, categorizing them into three groups: a) data transformations, b) specialized architectures, and c) regularization models. In this study, we focus on the first category: data transformations. Iqbal et al. [24] introduced a novel feature embedding technique Dynamic Weighted Tabular Method (DWTM), which dynamically uses feature weights based on their strength of the correlations to the class labels during applying any CNN architecture to the tabular data. In their approach each feature in the observation vector is assigned space in the image canvas based on its corresponding weight. They use statistical techniques such as Pearson correlation, chi-square test to compute the weights of each feature. Their results show that DWTM usually outperforms the results of traditional ML algorithms. Medeiros et al. [25] have provided comparative analysis of tabular data into image for classification. They conclude that transforming tabular data into images to leverage the power of CNN has the potential to increase the model performance by the additional 2D spatial information that can be processed by CNN. Their study highlights the potential benefits and limitations of using image-based DL models for tabular data. Kulkarni [3] proposed a method to map tabular data into images. They mapped feature values and ratios of the feature values as rectangular shapes in the image canvas. They used the model to classify tabular data. Li et al. [26] provide a survey on Graph Neural Networks (GNNs) for tabular data. The survey highlights a critical gap in deep neural tabular data learning methods: the underrepresentation of latent correlations amongst data instances and feature values.

## III. PROPOSED FRAMEWORK

The framework for the proposed Fuzzy Convolution Neural Network (FCNN) is shown in Fig. 1. We analyzed six tabular data sets using the proposed framework. The columns represent the features and rows represent entities. The last column in the training set data represents class labels. The first module in the proposed framework is a fuzzifier block, which converts feature values into the corresponding fuzzy memberships. The second module is a converter that maps fuzzy memberships onto the image canvas. During the training phase images are stored in Datamart. The last block is CNN, which is trained using the images in Datamart. During decision making phase, an unknown input feature vector is fuzzified. The fuzzified vector is converted into an image which is classified with the trained FCNN.

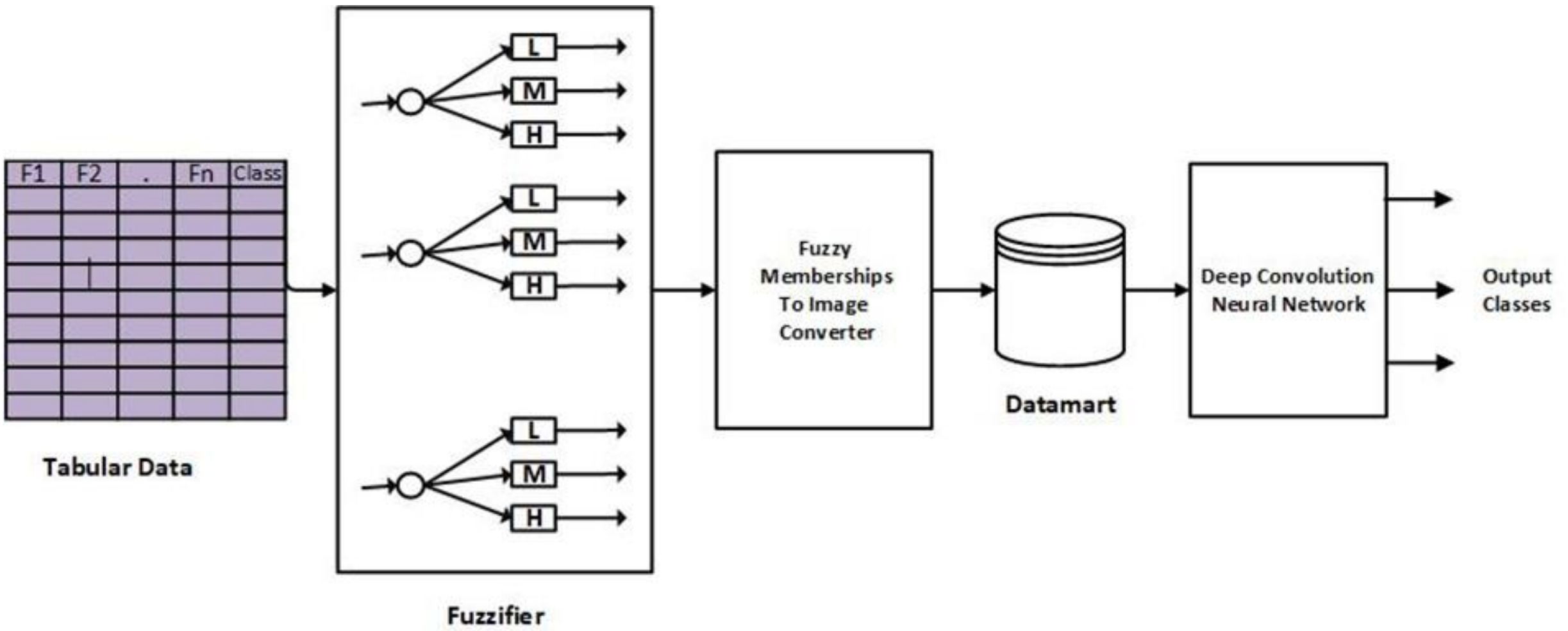

**FIGURE 1.** **Framework for fuzzy convolution neural network (FCNN)**

The trapezoidal and π-shaped fuzzy membership functions are shown in Fig. 2 and Fig. 3, respectively. The trapezoidal membership functions are given in (1).

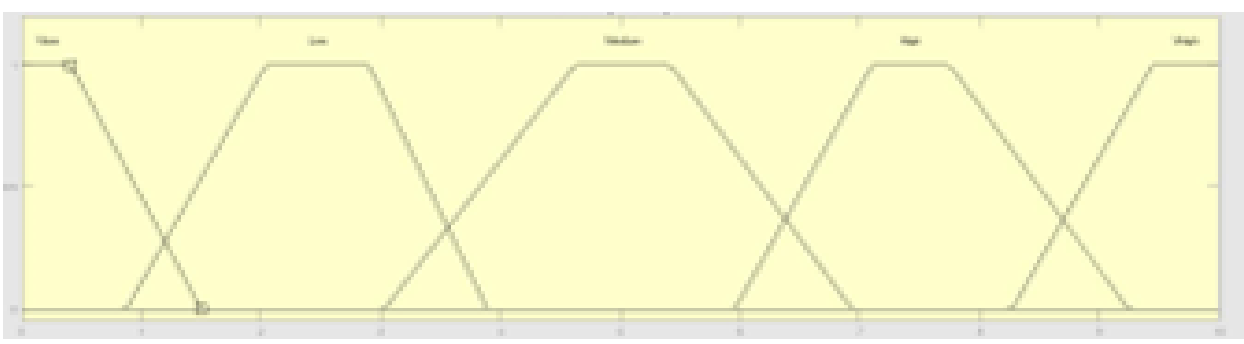

**FIGURE 2.** **Trapezoidal membership functions**

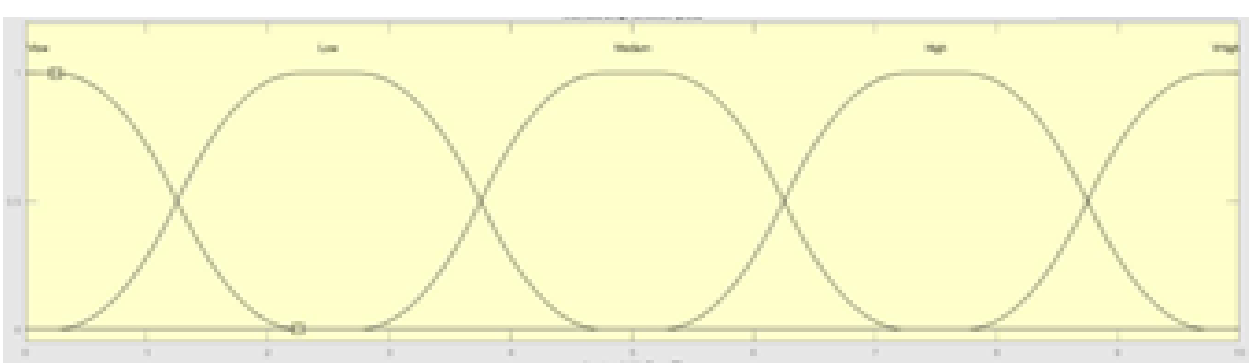

**FIGURE 3.** **π-shaped membership functions**

$$f(x;a,b,c,d)=\begin{cases} 0 & for \quad x<a \\ \dfrac{x-a}{b-a} & for \quad a\le x<b \\ 1 & for \quad b\le x<c \\ \dfrac{d-x}{d-c} & for \quad c\le x<d \\ 0 & for \quad d>0 \end{cases} \tag{1}$$

$$f(x;b,c)=\begin{cases} S\left(x;c-b,\dfrac{c-b}{2},c\right) & \text{for } \; x\le c \\ 1-S\left(x;c,\dfrac{c+b}{2},c+b\right) & \text{for } \; x>c \end{cases} \tag{2}$$

Where $a$, $b$, $c$, $d$ are constant that define the fuzzy membership function. The π-shaped membership functions are given by (2) and $S\left(x,a,b,c\right)$ represents a membership function which is defined in (3) [27].

$$S(x;a,b,c)=\begin{cases} 0 & \text{for } x<a \\ \dfrac{2\left(x-a\right)^2}{\left(c-a\right)^2} & \text{for } a\le x<b \\ 1-\dfrac{2\left(x-c\right)^2}{\left(c-a\right)^2} & \text{for } b\le x\le c \\ 1 & \text{for } x>c \end{cases} \tag{3}$$

In (3), $a$, $b$, and $c$ are the parameters that are adjusted to fit the desired membership function. The parameter $b$ is the half width of the curve at the crossover point. The triangular membership functions are defined by three parameters $a$, $b$, and $c$ as shown in (4).

$$f(x;a,b,c)=\begin{cases} 0 & for \quad x<a \\ \dfrac{x-a}{b-a} & for \quad a\le x<b \\ \dfrac{c-x}{c-b} & for \quad b\le x\le c \\ 0 & for \quad c>x \end{cases} \tag{4}$$

The Gaussian membership functions are defined by mean and standard deviation as shown in (5).

$$f(x;\sigma,c) = \exp\left[\frac{-(x-c)^2}{2\sigma^2}\right] \qquad (5)$$

Where c represents the mean value and σ represents the standard deviation.

Both trapezoidal and π-shaped membership functions (MFs) offer advantages over triangular and Gaussian MFs, particularly in terms of computational efficiency, robustness to noise, and flexibility. The flat top of these functions allows for a range of values to have full membership, which is beneficial when precise membership values are less critical. The flat top region can enhance the system's robustness to small variations or noise in the input, as a range of input values will share the same membership degree. Trapezoidal MFs are computationally less intensive compared to Gaussian MFs. They are straightforward to define and implement, requiring only four parameters. Their simplicity and practical applicability in fuzzy inference systems make trapezoidal MFs especially advantageous. On the other hand, π-shaped MFs provide a balance between smooth transitions and robustness, combining the characteristics of both trapezoidal and Gaussian functions. They are smooth like Gaussian functions but also feature a flat top like trapezoidal functions. These features make π-shaped MFs suitable for a wide range of fuzzy logic applications where both smooth transitions and robustness are desired. Triangular MFs, with their linear transitions, can result in more abrupt changes, especially if the input value is close to the peak of the triangle. Managing overlapping regions can be challenging, as the membership value changes linearly and abruptly at the boundaries, making the system more sensitive to input variations. Any change in input directly affects the membership value due to the linear nature of the function. The advantage of triangular MFs lies in their ease of implementation. Gaussian MFs, which require exponential calculations, can be computationally more intensive. Although they provide smooth curves, adjusting their shape precisely can be less intuitive since changes in parameters affect both the spread and the height of the curve simultaneously. In situations where the data distribution is normal with known mean and variance values, Gaussian MFs can represent the system more accurately. In our FCNN system implementation, we have chosen trapezoidal MFs.

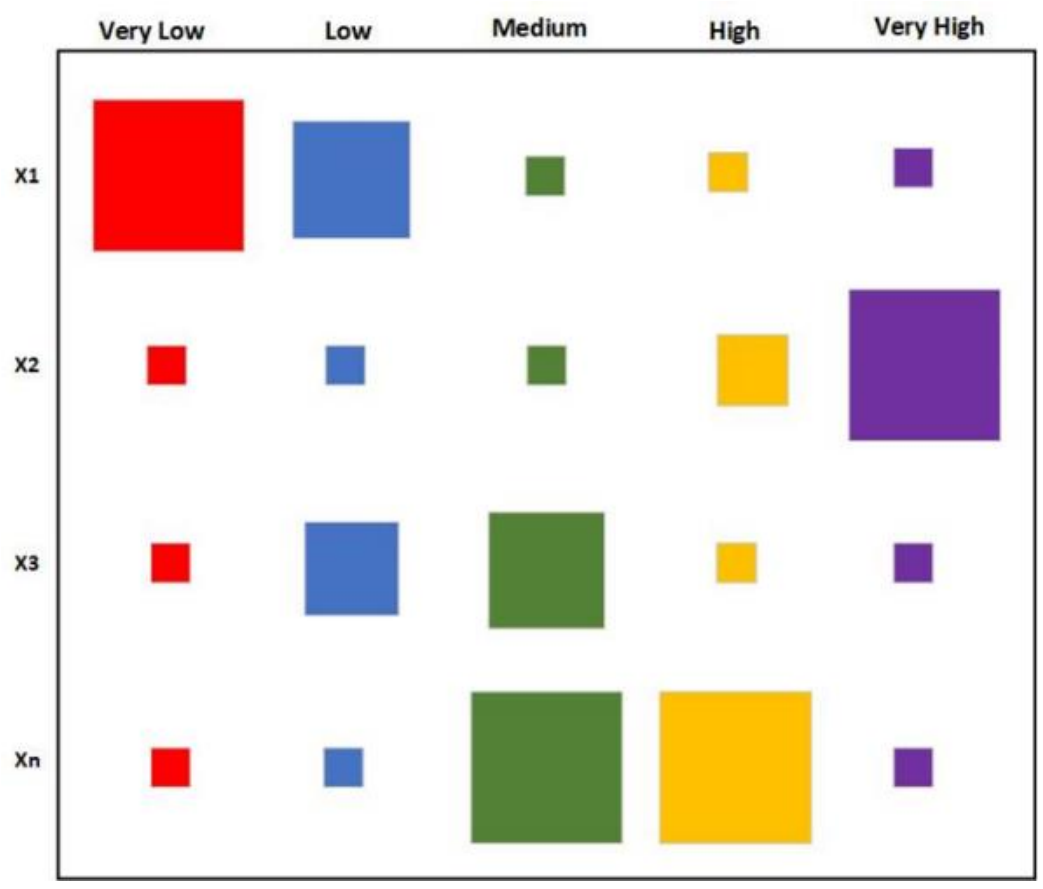

**FIGURE 4. Sample mapped image.**

The fuzzified feature vectors are mapped into images, which are saved in Datamart in folders that are labeled with class names. The mapped sample image is shown in Fig. 4. The shapes within the resulting mapped image symbolize the fuzzy membership values. The number of columns of shapes is equal to the number of term sets, while the number of rows is equal to the number of features in the observation vector. The number of the squares in the output image is equal to $n_f \times n_{term}$ where $n_f$ is the number of features and $n_{term}$ is the number of term sets. The area of each square in the mapped image is proportional to the corresponding fuzzy membership value. We analyzed six datasets that contain two features for each observation.

The last module implements DCNN model. The DCNN models are trained with the images stored in the images stored in the Datamart. Convolution layers extract features from the input image. Inputs are convolved with learned weights to compute feature maps and results are sent through a nonlinear activation function. The convolution layer is followed by a pooling layer. All neurons within a feature map have equal weights, however, different feature maps within the same. convolution layers have different weights. [11]. The output of the kth feature map $Y_k$ is given by (6)

$$Y_k = f\left(W_k * x\right) \qquad (6)$$

Where $x$ denotes the input image, $W_k$ is the convolution filter, and the '*'sign represents the 2D convolution operator. The purpose of the pooling layer is to reduce the spatial resolution and extract invariant features. The output of a pooling layer if given by (7).

$$Y_{kij} = \max_{(p,q)\in R_{ij}}\left(X_{kpq}\right) \qquad (7)$$

Where $X_{kpq}$ denotes elements at location ($p$, $q$) contained by the pooling region $R_{ij}$. We used two DCNN models in our analysis, the Alex Net and Resnet-50. Alex Net is seminal CNN architecture that significantly contributed to the advancement of deep learning in computer vision tasks. AlexNet consists of eight layers: five convolution layers followed by max-pooling layers, and three fully connected layers. The ReLu activation function is used throughout the network, and dropout regularization is applied to prevent overfitting. The network has the image input size of 227-by-227. The network maximizes the multinomial logistic regression objective function. Resnet-50 is DCNN, which is a variant of Resnet architecture. It is one of the most popular and influential deep learning models used for image classification and related tasks. It uses residual connections that allow the network to learn a set of residual functions that map the input to desired output. These connections enable the network to learn without suffering from vanishing gradients. It has fifty layers. The architecture is divided into four parts: convolution layers, the identity block, the convolution block, and fully connected layers. It introduced the concept of residual connections, which are shortcut connections that skip one or more layers. These connections allow gradients to flow more easily during training, mitigating the vanishing gradient problem and enabling training of very deep networks. ResNet50 employes a bottleneck architecture, which reduces the computational cost of the network by using 1x1, 3x3, 1x1 convolutions in sequence. Resnet50 is often used as a pre-trained model for transfer learning. The pre-trained model can be fine-tuned on a smaller dataset for a specific task [28]. In our research work we used both AlexNet and Resnet50 to train and classify images that were generated from fuzzified feature vectors.

## IV. EXPERIMENT AND RESULTS

To validate our approach, we generated six artificial noisy non-linearly separable datasets: Half Kernel, Two Spirals, Clusters in Cluster, Crescent and Moon, Corners, and Outliers. Scatter plots for the datasets are displayed in Fig. 5. Each dataset comprises two attributes and 400 samples. All datasets, except for the Corners dataset, consist of samples from two classes, with 200 samples per class represented by \

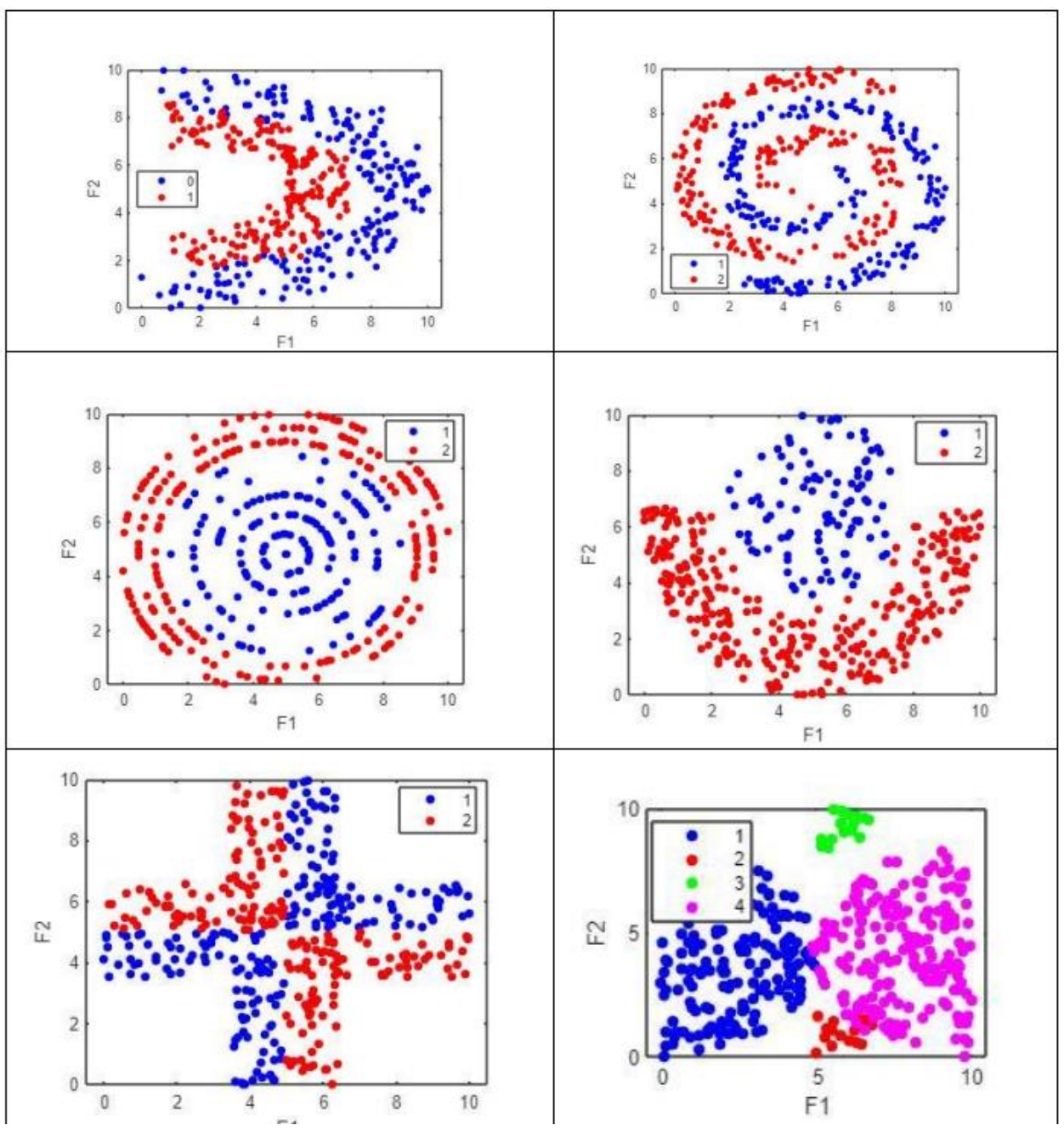

**FIGURE 5. Scatter plots for Half Kernel, Spirals, Cluster in Cluster, Crescent, Corners, Outliers datasets.**

blue and red dots. The Corners dataset contains samples from four classes. Two datasets, Half Kernel and Corners, exhibit overlapping samples in the feature space. The datasets were generated using a MATLAB script [29]. The Outliers dataset comprised of 200 samples from four classes. The data set contains overlapping samples in the feature space. As an illustration, the results from analysis of the Two Spiral dataset are presented below. The parallel coordinates plot for the Two Spirals dataset is shown in Fig. 6.

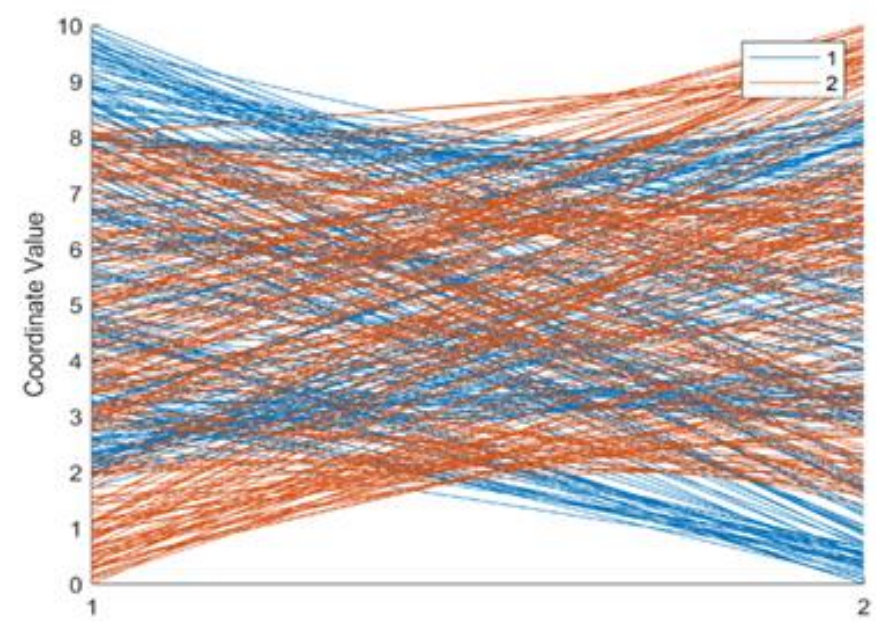

**FIGURE 6. Parallel coordinates for Two Spiral dataset.**

The dataset was split for training and testing. 70 percent of randomly selected samples were selected for training and the remaining 30 percent were used for evaluating the models. The decision tree that was generated to the Two Spiral data set is shown in Fig. 7. The confusion matrix is shown in Fig. 8. The decision tree classifier was able to classify dataset with 92.5 percent accuracy. Fig. 9 shows the ROC curve obtained with the DT classifier. The SVM and Bayes' classifier were able to classify the dataset with 65 percent accuracy, and with the RF we got the accuracy of 95 percent.

We also classified the Two Spiral dataset using the fuzzy neural network (FNN) shown in Fig. 10. The fuzzy membership values were used as the input for the neural network. FNN model consists of two modules. The first module is a fuzzifier module that maps feature values into fuzzy membership functions. We have used five trapezoidal membership functions that represent five term sets. The neural network has 10 input units that represent fuzzy membership values for the two features. The hidden layer has ten units, and the output unit is with two units that represent two classes. The same dataset was classified by FNN model with the accuracy of 86 percent. The learning curve for the FNN model is shown in Fig 11.

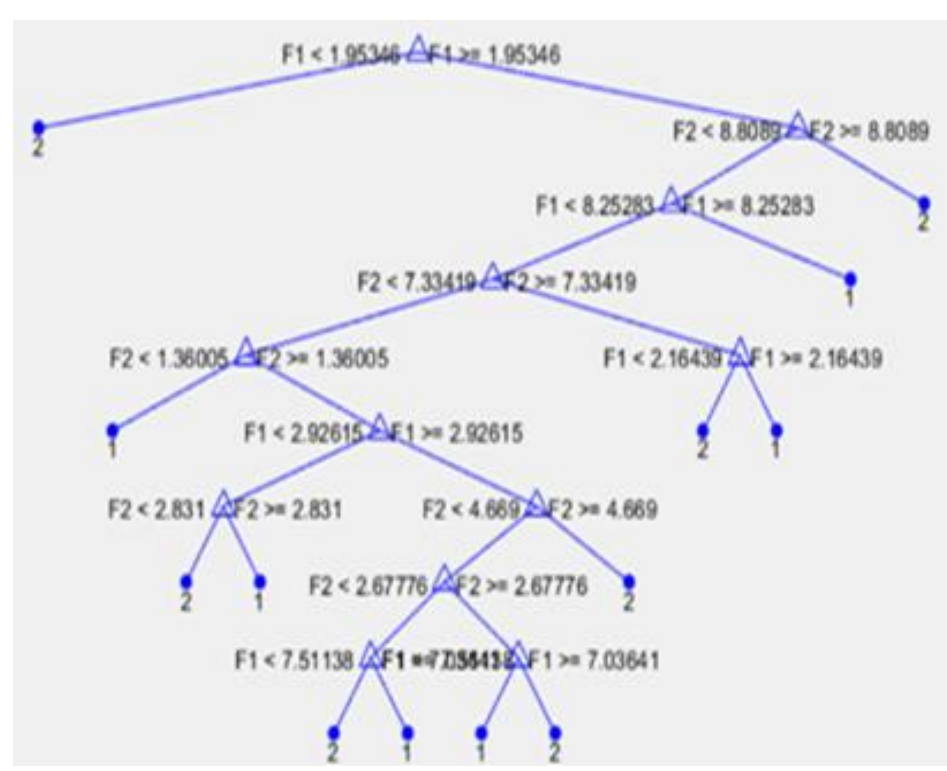

**FIGURE 7. Decision tree for Two Spiral dataset**

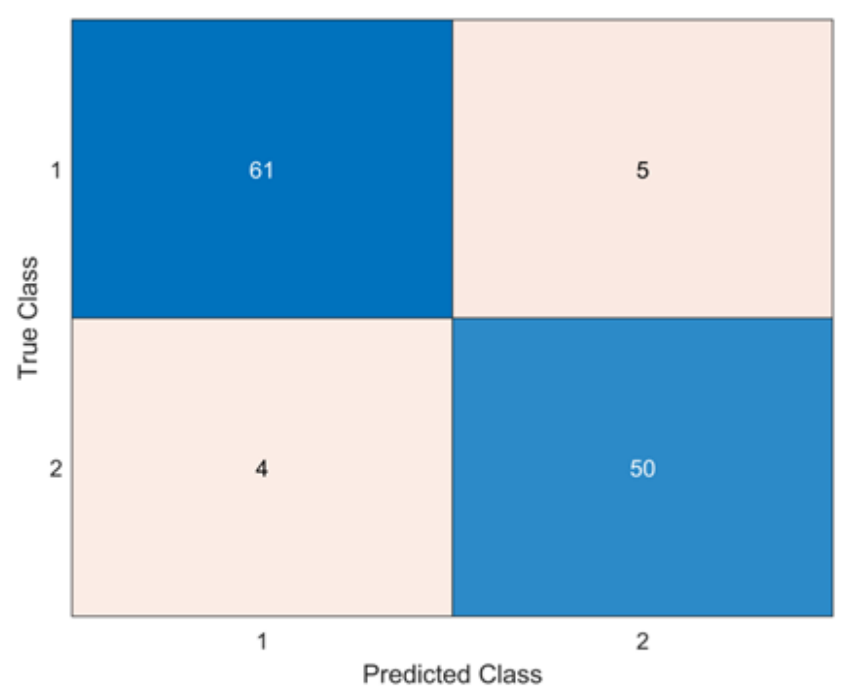

FIGURE 8. Confusion matrix for decision tree for Two Spiral dataset.

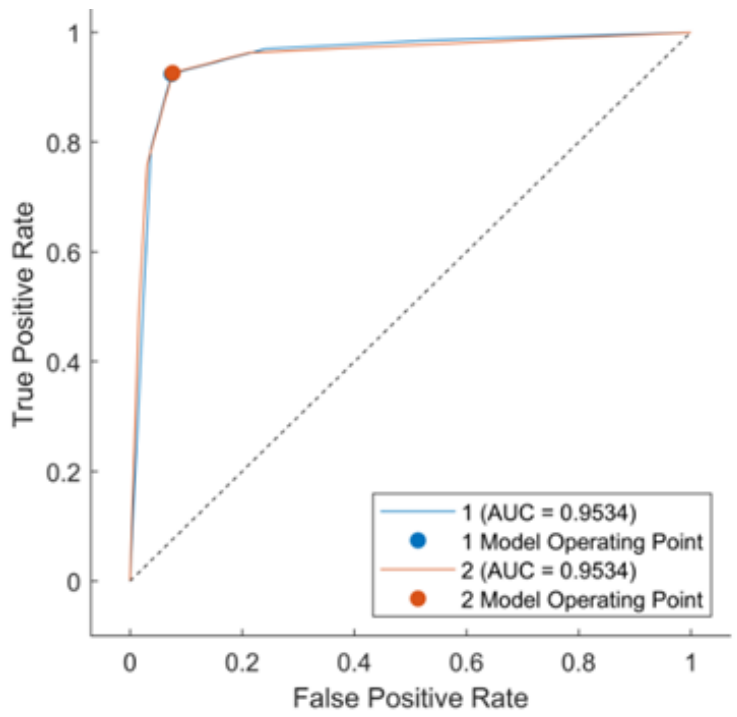

FIGURE 9. ROC curve for decision tree for Two Spiral dataset.

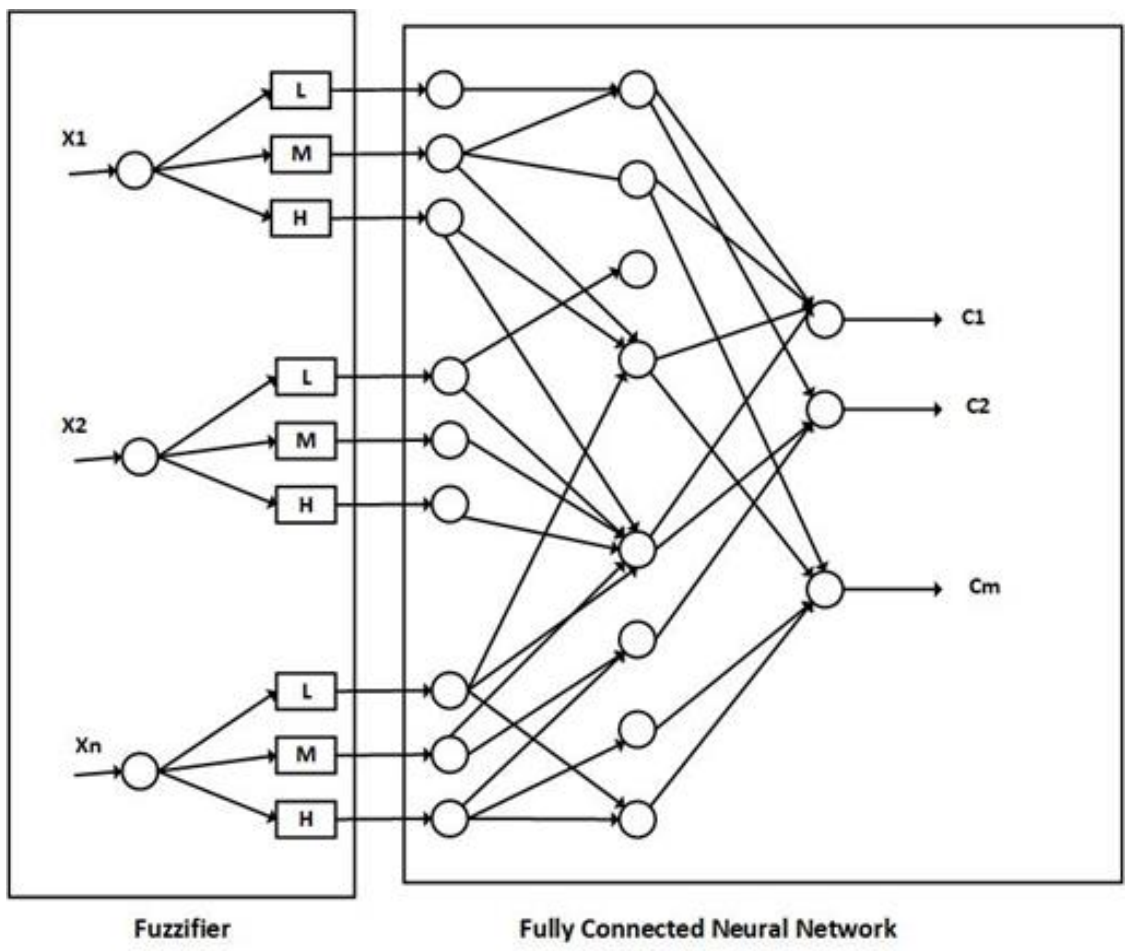

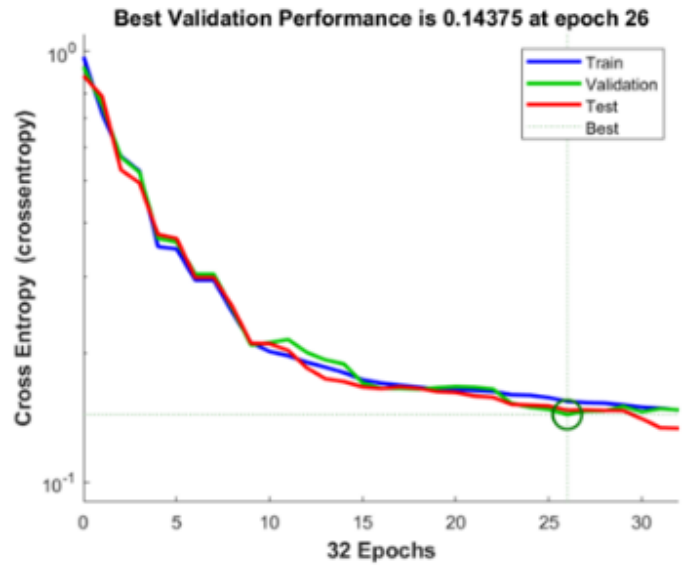

FIGURE 11. Error curve for FNN learning

The Two Spiral data set was analyzed by the proposed FCNN model. We developed software using MATLAB script to map fuzzy membership values to images. Each fuzzified feature vector was mapped to an image. The mapped images were stores in respective class folders in Datamart. We implemented two FCNN models, one with Alex Net and the other with Resnet 50 using MATLAB script. The input image size for Alex Net was 227 x 277 x 3, and the input image size for Resnet 50 was 512 x 512 x 3. The number of output units for both models was equal to the number of classes. The training progress plots for Alex Net and Resnet50 are shown in Fig 12 and Fig. 13, respectively. For training two FDNN models 70 percent of randomly chosen images were used for training and the remaining 30 percent were used for testing. Both FCNNs were able to classify images in the testing set with 100 percent accuracy. and Resnet-50. We implemented and executed FCNN models with both Alex Net and Resnet-50 on a desktop with a Pentium dual processor. The execution time can be decreased by executing the script on a workstation with a GPU. The training process for Alex Net took about 4 min and 50 sec for each data set for 28 iterations, while learning process for Resnet-50 took about 78 min for each data set for 72 iterations. Some sample classified images with class labels are shown in Fig. 14. In this example, the dataset consists of two features. While mapping features, we mapped each feature twice. The first two rows of shapes represent the first feature, and the last two rows of shapes represent the second feature. The ROC Curves for both FCNN models are shown in Fig. 15 and 16. All six datasets were classified using ML models that include decision tree (DT), support vector machine (SVM), Bayes' classifier, Random Forest (RF) and fuzzy neural network (FNN). The classification accuracy obtained by with these classifies for all six datasets is shown in Table I. The classification accuracy for FCNN models with AlexNet and Resnet-50 was the same for all datasets.

TABLE I. CLASSIFICATION ACCURACY

| | Half Kernel | Two Spirals | Cluster-in-Cluster | Crescent Moon | Corners | Outliers |
|---|---|---|---|---|---|---|
| Decision Tree | 95.00 | 90.83 | 94.17 | 97.50 | 98.33 | 98.33 |
| Support Vector Machine | 66.67 | 65.00 | 56.67 | 85.00 | 58.33 | 99.17 |
| Bayes' Classifier | 84.17 | 65.00 | 87.50 | 88.30 | 50.00 | 99.17 |
| Random Forest | 98.33 | 95.00 | 95.83 | 97.5 | 71.67 | 44.17 |
| Fuzzy Neural Network | 91.5 | 86.6 | 100.00 | 98.50 | 53.0 | 95.2 |
| Fuzzy Convolution Neural Network | **99.19** | **100.00** | **100.00** | **100.00** | **100.00** | **99.17** |

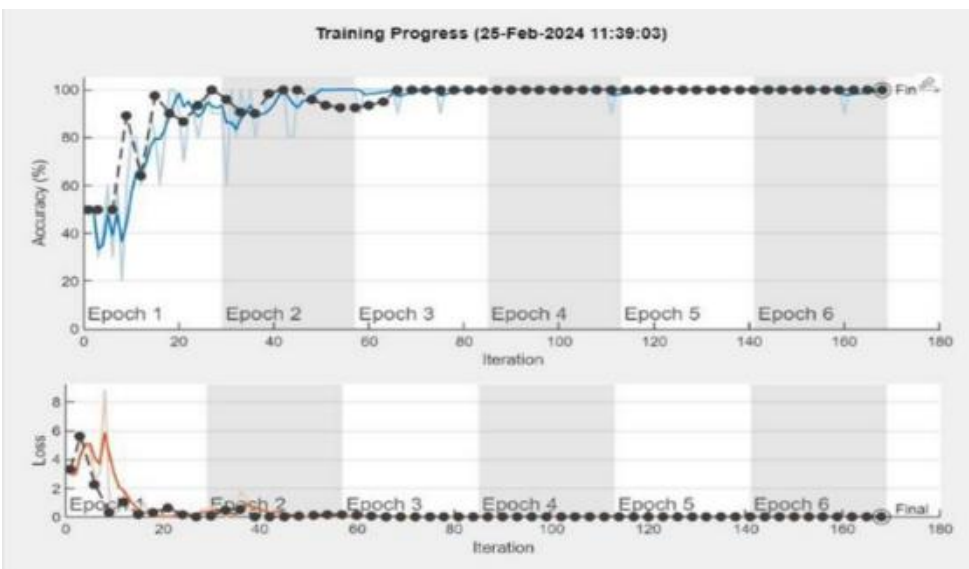

FIGURE 12. Training progress plot for FCNN (Alex Net)

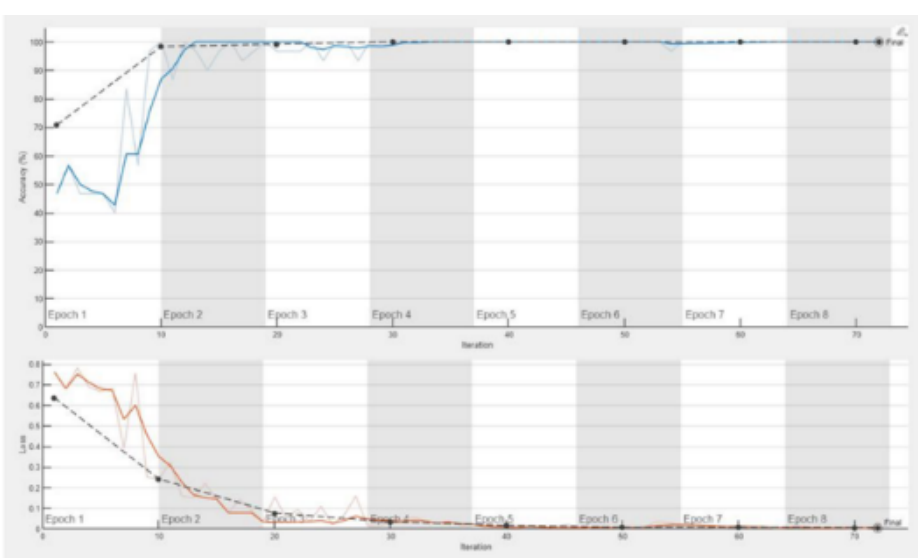

FIGURE 13. Training progress plot for FCNN (Resnet-50)

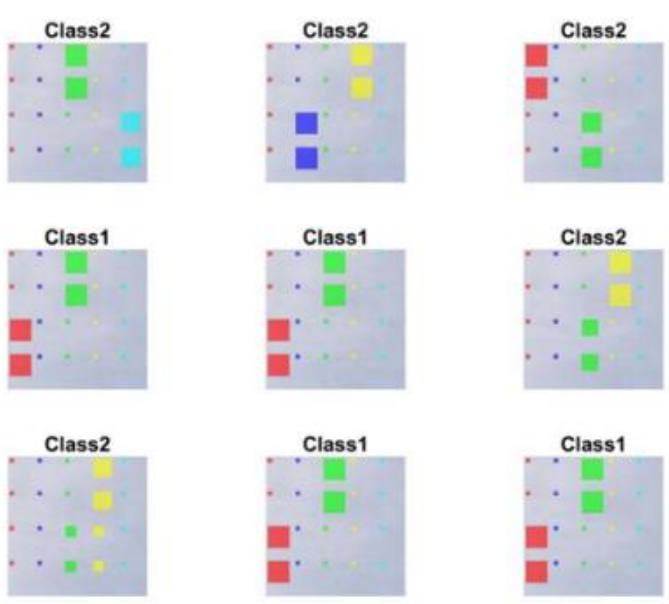

FIGURE 14. Classified images with labels (FCNN-Resnet-50)

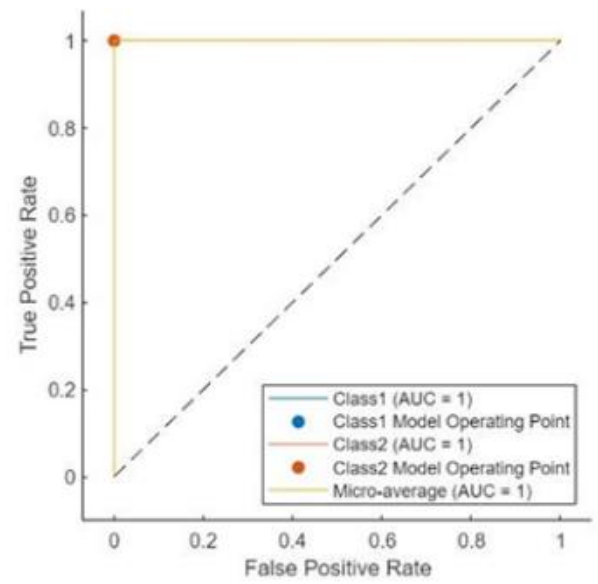

FIGURE 15. ROC curve for Two Spiral dataset (Alex Net)

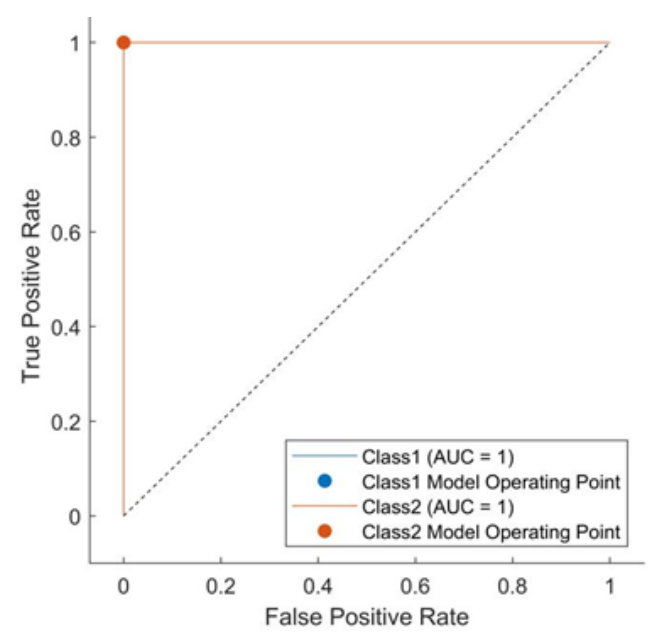

FIGURE 16. ROC curve for Two Spiral dataset (Resnet-50)

## V. CONCLUSIONS

In this paper, we present a novel framework called FCNN for classifying tabular data. We developed software using MATLAB scripts to map features to corresponding fuzzy membership values and to convert fuzzified vectors into images. Additionally, we implemented AlexNet and ResNet-50 using the MATLAB Deep Learning Toolbox. To evaluate the proposed approach, we generated six complex noisy datasets and analyzed them using various ML algorithms: decision trees, support vector machines, Bayes' classifiers, Random Forests, and fuzzy neural networks. The six datasets were also classified using the proposed FCNN model. It can be observed from Table I that FCNN model performs as well as or better than state-of-the-art ML algorithms, suggesting that FCNN provides a viable alternative for classifying tabular data. The limitation of the proposed approach is the number of features and term sets.

The number of shapes in the mapped image is proportional to the number of features and term sets. For a finite image size, the number of shapes that can be mapped onto the image canvas is limited. Therefore, the approach is suitable for datasets with a small number of features. The future work includes a) It is possible to directly feed images to the DCNN and eliminate Datamart, b) Experimenting with shapes having different morphological properties, such as circular, rectangular, hexagonal, or triangular to generate mapped images. c) In this research work, we have used trapezoidal fuzzy membership function. We would like to try other membership functions such as Gaussian and triangular and evaluate the classification accuracy. d) In our current research, we have utilized AlexNet and ResNet-50 DCNN models. We would like to analyze data using other DCNNs such as VGG-16 and GoogleNet and deploy the FCNN model to real-life applications.

## ACKNOWLEDGMENT

The author would like to express sincere thanks to the anonymous reviewers for their insightful comments and constructive suggestions, which have significantly contributed to the improvement of this manuscript.

## REFERENCES

[1] V. Borisov, T. Leemann, K. Seßler, J. Haug, M. Pawelczyk, and G. Kasneci. “Deep Neural Networks and Tabular Data: A Survey,” *IEEE Transactions on Neural Networks and Learning Systems*, 2022. (accepted for publication) https://doi.org/10.48550/arXiv.2110.01889

[2] V. Martin, “Convolutional Neural Networks on Tabular Datasets (Part 1)”, 2021. https://medium.com/spikelab/convolutional-neural-networks-on-tabular-datasets-part-1-4abdd67795b6

[3] A. D. Kulkarni, “Multispectral Image Analysis Using Convolution Neural Networks,” *International Journal of Advanced Computer Science and Applications*, vol. 14, no. 10, 2023, pp. 13-19. doi: 10.14569/IJACSA.2023.0141002

[4] R. O. Duda, P. E. Hart, and D. G. Stork, *Pattern classification*. John Wiley & Sons, New York, 2001.

[5] T. Mitchell, *Machine Learning*, WCB/McGraw-Hill, Boston, MA, 1997, pp. 52-80.

[6] Y. Song, Y. Lu, “Decision tree methods: applications for classification and prediction*,” Shanghai Arch Psychiatry*. 2015 Apr 25;27(2):130-5. doi: 10.11919/j.issn.1002-0829.215044. PMID: 26120265; PMCID: PMC4466856.

[7] I. Breiman. “Random Forests.” *Machine Learning*, vol 45, 2001, pp. 5–32. https://doi.org/10.1023/A:1010933404324

[8] V. Vapnik, and S. Kotz, (2006). Estimation of Dependences Based on Empirical Data. 2006 doi: 10.2307/2988246.

[9] B. Mehlig, *Machine learning with neural networks: an introduction for scientists and engineers*. Cambridge University Press, 2021.

[10] A. Krizhevsky, I. Sutskever, G. Hinton, “ImageNet classification with deep convolutional neural networks.” *Adv Neural Inf Process Syst.* 2012. https://doi.org/10.1061/(ASCE)GT.1943-5606.0001284

[11] Y. LeCun, Y. Bengin, and G. Hinton, “Deep learning,” *Nature*, vol. 521, 2015, pp. 436-444,

[12] W. Rawat, and Z. Wang, “Deep convolution neural networks for image classification: A comprehensive review,” *Neural Computation*, vol. 29, 2017, pp. 2352-2449.

[13] S. Zhang, L. Yaq, A. Sun, Y. Tay, “Deep Learning based Recommender System: A Survey and New Perspectives,” *ACM Computing Surveys*, 2018, vol. 1, no. 1, pp. 1-35.

[14] Q. Abbas, M. Ibrahim, M. Jaffar (2019) A comprehensive review of recent advances in deep vision systems. *Artificial Intelligence Review*, 2019, vol. 52, pp 39–76. https://doi.org/10.1007/s10462-018-9633-3

[15] A. Khan, A. Sohail, U. Zahoora, A. S. Qureshi, “A Survey of the Recent Architectures of Deep Convolutional Neural Networks,” *Artificial Intelligence Review*, 2020, vol. 53, pp. 5455–5516. https://doi.org/10.1007/s10462-020-09825-6.

[16] N. Talpur, S. J. Abdulkadir, H. Alhussian, M. Hilmi Hasan, N. Aziz, A. Bamhdi. “Deep Neuro-Fuzzy System application trends, challenges, and future perspectives: a systematic survey,” *Artificial Intelligence Review*, 2023, vol. 56, pp 865–913. https://doi.org/10.1007/s10462-022-10188-3

[17] A. Sarabakha, E. Kayacan E, “Online deep fuzzy learning for control of nonlinear systems using expert knowledge*,” IEEE Transactions on Fuzzy Systems*, vol. 28, no. 7, 2019, pp. 1492–1503. https://doi.org/10.1109/TFUZZ.2019.2936787

[18] R. Das, S. Sen, and U. Maulik, “Survey on fuzzy deep neural networks.” *ACM Comput. Survey*, vol. 53, no. 3, May 2020. https://dl.acm.org/doi/abs/10.1145/3369798

[19] B. Sun, L. Yang, P. Dong, W. Zhang, J. Dong, and C. Young. ‘‘Super characters: A conversion from sentiment classification to image classification,” 2018, arXiv:1810.07653.

[20] A. Sharma, E. Vans, D. Shigemizu, K. A. Boroevich, and T. Tsunoda, “Deep Insight: A methodology to transform a non-image data to an image for convolution neural network architecture.” *Nature Sci Rep* vol. 9, 2019, pp. 11399...

[21] Zhu, Y., Brettin, T., Xia, F. et al.,” Converting tabular data into images for deep learning with convolutional neural networks.” *Nature Sci Rep*, vol. 11, 2021, pp.11325 https://doi.org/10.1038/s41598-021-90923-y

[22] L. Du, et al, “TabularNet: A Neural Network Architecture for Understanding Semantic Structures of Tabular Data,” *KDD* ’21, August 14–18, 2021, Virtual Event, Singapore, pp 322-331.

[23] S. O. Arik and T. Pfister, “TabNet: Attentive Interpretable Tabular Learning.,” *Proceedings of the AAAI Conference on Artificial Intelligence*, vol. 35, no. 8, 2021, pp. 6679-6687. https://doi.org/10.1609/aaai.v35i8.16826

[24] Md. I. Iqbal et al.: Dynamic “Weighted Tabular Method for Convolutional Neural Networks Convolutional Neural Networks,” *IEEE Access*, vol 10,2022, pp 134183- 134198

[25] N. I. Medeiros, N. S. Rogerio da Silva, P. T. Endo, “A comparative analysis of converters of tabular data into image for the classification of Arboviruses using Convolutional Neural Networks,” *PLoS ONE* 2023, vol.18, no. 12: 2023, e0295598. https://doi.org/10.1371/journal.pone.0295598

[26] C.-T. Li, Y.-C. Tsai, C.-Y. Chen, J. C. Liao. “Graph Neural Networks for Tabular Data Learning: A Survey with Taxonomy & Directions,” 2024. https://github.com/Roytsai27/awesome-GNN4TDL

[27] A. D. Kulkarni, *Computer Vision, and Fuzzy Neural Systems*. 2001, Prentice Hall, Upper Saddle River, NJ.

[28] K. He, X. Zhang, S. Ren, and J. Sun. ‘‘Deep residual learning for image recognition,’ in *Proceeding of. IEEE Conf. Comput. Vis. Pattern Recognit. (CVPR)*, Jun. 2016, pp. 770–778.

[29] J. Kools, “6 functions for generating artificial datasets”, https://www.mathworks.com/matlabcentral/fileexchange/41459-6-functions-for-generating-artificial-datasets, MATLAB Central File Exchange. Retrieved March 7, 2024.

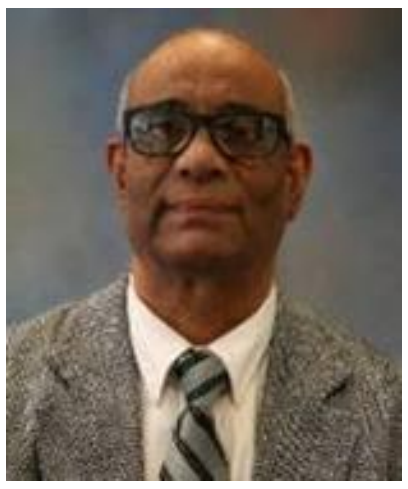

ARUN KULKARNI obtained M. Tech. and Ph.D. degrees from the Indian Institute of Technology, Mumbai, and was a post-doctoral fellow at Virginia Tech. His areas of interest include machine learning, data mining, deep learning, and computer vision. He has more than eighty refereed papers to his credit and has authored two books. Currently, he is working as a Professor of Computer Science, with The University of Texas at Tyler. His awards include the Office of Naval Research (ONR) 2008, Senior Summer Faculty Fellowship award, 2005-2006 President’s Scholarly Achievement Award, 2001-2002 Chancellor's Council Outstanding Teaching award, 1997 NASA/ASEE Summer Faculty Fellowship award, and the 1984 Fulbright Fellowship award. He has been listed in who's who in America.